\documentclass[sigconf]{acmart}
\AtBeginDocument{%
  }

\copyrightyear{2026}
\acmYear{2026}
\setcopyright{cc}
\setcctype{by}
\acmConference[MM '26]{Proceedings of the 34th ACM International Conference on Multimedia}{November 10--14, 2026}{Rio de Janeiro, Brazil}
\acmBooktitle{Proceedings of the 34th ACM International Conference on Multimedia (MM '26), November 10--14, 2026, Rio de Janeiro, Brazil}
\acmDOI{10.1145/3767308.3835797}
\acmISBN{979-8-4007-2213-4/2026/11}

\usepackage{booktabs}
\usepackage{multirow}
\usepackage{xcolor}
\usepackage{colortbl}

\definecolor{rowgray}{gray}{0.93}
\definecolor{posgreen}{RGB}{0,140,0}
\definecolor{negred}{RGB}{200,0,0}

\newcommand{\dpos}[1]{\textcolor{posgreen}{\scriptsize(+#1)}}
\newcommand{\dneg}[1]{\textcolor{negred}{\scriptsize(-#1)}}

\begin{document}

\title{Rethinking Where to Edit: Task-Aware Localization for Instruction-Based Image Editing}

\author{Jingxuan He}
\email{jihe0215@uni.sydney.edu.au}
\affiliation{%
  \institution{The University of Sydney}
  \city{Sydney}
  \country{Australia}
}

\author{Xiyu Wang}
\email{xiyuwang.usyd@gmail.com}
\affiliation{%
  \institution{The University of Sydney}
  \city{Sydney}
  \country{Australia}
}

\author{Mengyu Zheng}
\email{myzheng@bjtu.edu.cn}
\affiliation{%
  \institution{Beijing Jiaotong University}
  \city{Beijing}
  \country{China}
}

\author{Xiangyu Zeng}
\email{xzen0351@uni.sydney.edu.au}
\affiliation{%
  \institution{The University of Sydney}
  \city{Sydney}
  \country{Australia}
}

\author{Yunke Wang}
\email{yunke.wang@sydney.edu.au}
\affiliation{%
  \institution{The University of Sydney}
  \city{Sydney}
  \country{Australia}
}

\author{Chang Xu}
\correspondingauthor
\email{c.xu@sydney.edu.au}
\affiliation{%
  \institution{The University of Sydney}
  \city{Sydney}
  \country{Australia}
}

\renewcommand{\shortauthors}{Jingxuan He et al.}

\begin{abstract}
Instruction-based image editing (IIE) aims to modify an image according to a natural language instruction. Despite recent advances in diffusion transformers, existing methods often introduce unintended changes to regions unrelated to the requested edit. We attribute this limitation to the absence of an explicit mechanism for edit localization. Different editing operations (e.g., subject addition, removal, and replacement) induce distinct spatial patterns, yet existing IIE models typically perform localization in a task-agnostic manner. To address this limitation, we propose a training-free, task-aware edit localization framework that exploits the intrinsic source and target image streams of IIE models. For each image stream, we construct feature centroids from attention-based edit cues, and then partition tokens into edit and non-edit regions based on feature similarity. Observing that effective localization is inherently task-dependent, we introduce a unified mask construction strategy that selectively leverages the source and target streams according to the editing task. We also provide a systematic analysis of our underlying insights and design choices. Extensive experiments on EdiVal-Bench demonstrate that our framework consistently improves content consistency in non-edit regions while maintaining strong instruction-following performance on top of Step1X-Edit and Qwen-Image-Edit.
\end{abstract}





\maketitle

\section{Introduction}
\label{sec:intro}

Recent advances in diffusion-based visual generation have significantly enhanced the quality of synthesized images~\cite{ho2020denoising, song2020denoising, wang2023learning, rombach2022high, peebles2023scalable, esser2024scaling}.
Beyond pure synthesis, these developments have also positioned image editing as a critical capability, as many real-world applications necessitate modifying existing content rather than generating new images from scratch.
Among various image editing paradigms, instruction-based image editing (IIE)~\cite{brooks2023instructpix2pix, zhang2023magicbrush, zhao2024ultraedit, zhang2025enabling, liu2025step1x, wu2025qwen} is particularly appealing. By enabling users to specify desired modifications through natural language, it offers a more intuitive and flexible interface for sophisticated image manipulation.

\begin{figure}
    \centering
    \includegraphics[width=0.95\linewidth]{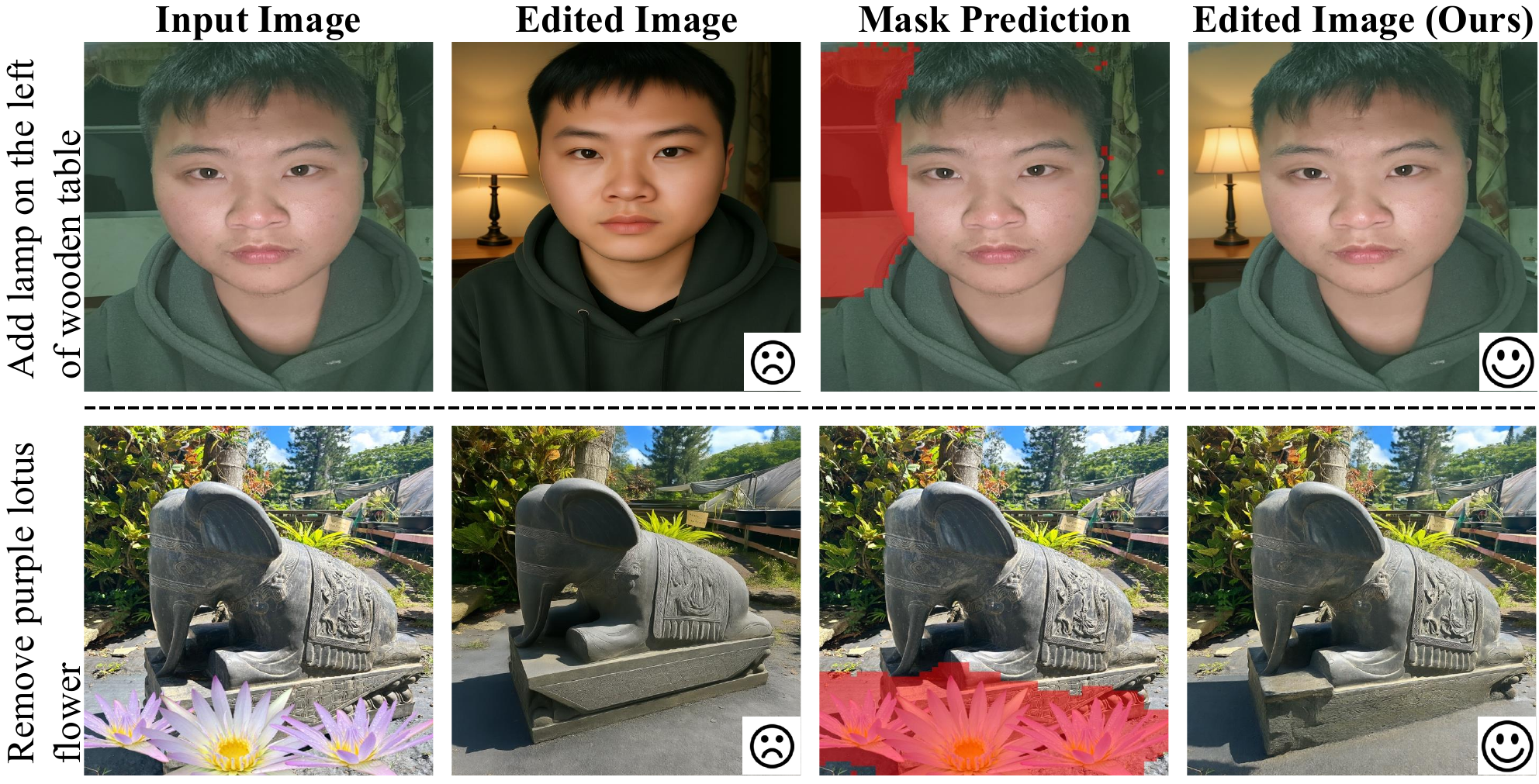}
    \caption{\small Motivation of our work. The base image editing model~\cite{wu2025qwen} produces plausible results, it also introduces unintended changes, such as human beautification in the first example and displacement of the elephant in the second example. Our framework aims to preserve content outside the intended region while still following editing instructions.}
    \label{fig:teaser}
\end{figure}

Despite these advances, current IIE models often exhibit a critical limitation: they tend to modify regions beyond those specified by the instruction, introducing undesired alterations to background content, object appearance, or the overall scene structure.
This phenomenon of \emph{over-editing} undermines a fundamental principle of localized image manipulation: a successful edit should faithfully execute the requested modification while preserving the source content in regions irrelevant to the instruction.
However, existing models frequently fail to satisfy this criterion, instead exhibiting an undesirable trade-off in which improved instruction following comes at the expense of content consistency in non-target regions.

This limitation is particularly pronounced in modern IIE models~\cite{liu2025step1x, wu2025qwen} built upon diffusion transformers (DiTs)~\cite{peebles2023scalable}.
These models typically process a unified sequence of tokens, where the streams of text, source image, and target image interact through joint attention during denoising.
While this design provides powerful cross-modal modeling capacity, it lacks an explicit mechanism to localize and restrict \emph{where} the edit occurs.
Without such explicit regularization, the model is burdened with redundantly reconstructing non-target regions, which may result in unintended deviations from the original content.
For instance, as illustrated in Fig.~\ref{fig:teaser}, Qwen-Image-Edit~\cite{wu2025qwen} successfully executes subject addition and removal; however, it also modifies the tone of a human portrait and displaces the elephant in the scene, respectively.
From this perspective, the primary weakness of modern IIE models is not a lack of generative power, but the absence of an explicit mechanism of edit localization.

In this work, we address this challenge from the perspective of task-aware edit localization.
Rather than redesigning the generative backbone or introducing additional supervision, we investigate a more fundamental question: which spatial regions should evolve during denoising, and which should remain anchored to the source image?
Answering this is non-trivial, as edit semantics do not emerge uniformly across the streams of an IIE model.
Our key insight is that \emph{effective localization is inherently task-dependent}: object addition requires identifying regions of emergence in the target image stream, whereas object removal focuses on localizing content within the source image stream.
More complex operations, such as subject replacement, require jointly considering both disappearance in the source stream and emergence in the target stream.

Building on this insight, we propose a training-free, task-aware edit localization framework for instruction-based image editing.
Our method begins by decomposing and propagating attention activations in a stream-specific manner, resulting in coarse localization cues.
Considering that attention primarily reflects token-wise relevance, we treat these attentive cues as semantic indicators and derive prototypical centroids via masked average pooling.
The edit mask for each image stream is obtained by evaluating the feature similarity of each image token to these centroids.
Next, we construct the task-aware edit mask by combining stream-specific edit masks based on the essential requirement of the editing task.
Finally, the resulting edit mask is utilized to guide latent update during denoising, enabling semantically grounded editing with minimal unintended changes.

Beyond the proposed framework, we provide a systematic analysis of how edit semantics emerge within DiT-based image editing models. 
Through qualitative and quantitative evaluations, we show that deeper DiT features provide more robust localization signals than attention activations.
Importantly, our analysis reveals that an effective localization strategy is inherently task-specific, which fundamentally differs across subject addition, removal, and replacement.
Extensive experiments on EdiVal-Bench~\cite{chen2025edival} demonstrate that our approach enhances content consistency in non-edited regions while maintaining strong instruction-following performance when integrated with powerful image editing backbones.

Our contributions are summarized as follows:
\begin{itemize}
    \item We identify the absence of explicit edit localization as a primary cause of over-editing in modern IIE models, and frame the problem as enhancing non-edit region consistency without compromising instruction-following performance.
    \item We propose a training-free edit localization framework, which extracts stream-wise edit signals from semantically rich latent features and constructs task-aware edit masks by dynamically integrating these signals for accurate edit localization.
    \item We provide a systematic analysis of how edit semantics emerge in DiT-based image editing models, revealing that latent features offer more reliable localization signals than attention activations, and that effective localization strategies are inherently task-dependent.
    \item Extensive experiments on EdiVal-Bench demonstrate that our method consistently improves content consistency in non-edited regions while maintaining strong instruction-following performance.
\end{itemize}
\section{Related Work}

\subsection{Instruction-Based Image Editing}
Instruction-based image editing (IIE) modifies an image according to a natural language instruction, which requires effective alignment between visual features and textual semantics.
Among the earliest efforts, InstructPix2Pix~\cite{brooks2023instructpix2pix} adapts {Stable Diffusion}~\cite{rombach2022high} for IIE by fine-tuning it on a synthetic dataset.
Subsequent work focuses on improving data quality and instruction diversity.
For example, MagicBrush~\cite{zhang2023magicbrush} introduces human-annotated data to reduce noise and improve realism, while UltraEdit~\cite{zhao2024ultraedit} constructs a large-scale dataset with real-image anchors and region-level annotations to enhance editing performance.
Methods such as MGIE~\cite{fu2023guiding} and SmartEdit~\cite{huang2024smartedit} incorporate multimodal large language models (MLLMs) to enhance instruction comprehension and reasoning.
More recently, the field has witnessed the emergence of unified models for image generation and editing~\cite{xiao2025omnigen, wu2025omnigen2, han2024ace, mao2025ace++, pu2025lumina}.
In parallel, specialized image editing frameworks have also advanced rapidly.
For instance, Step1X-Edit~\cite{liu2025step1x} and Qwen-Image-Edit~\cite{wu2025qwen} adopt a similar dual-stream formulation, where source and target image tokens are jointly processed within a diffusion transformer.
Although these models demonstrate strong instruction-following performance, they often lack explicit control over the spatial extent of edits, which may introduce over-editing artifacts.

\subsection{Localized Image Editing}
Localized image editing confines transformations to specific regions for global consistency.
Existing approaches are broadly mask-based or mask-free.
Mask-based methods, including GLIDE~\cite{nichol2021glide}, {Blended Diffusion}~\cite{avrahami2022blended}, DiffEdit~\cite{couairon2022diffedit}, and InstructEdit~\cite{wang2023instructedit}, use user-provided or externally generated masks to constrain generation.
Mask-free methods instead manipulate internal mechanisms of diffusion models.
{Prompt-to-Prompt}~\cite{hertz2022prompt} controls cross-attention to preserve spatial layout, while MasaCtrl~\cite{cao2023masactrl} introduces mutual self-attention to transfer local content and textures from the source image.
LIME~\cite{simsar2025lime} derives segments from intermediate features and regularizes attention for localization.
Unlike these methods, we focus on modern dual-stream editing models and show that edit semantics emerge differently across streams and tasks.
This leads to a task-aware localization strategy that improves non-edit region consistency without sacrificing instruction-following performance.
\section{Preliminaries: Image Editing Models}
\label{preliminaries}

\begin{figure*}[htbp]
    \centering
    \includegraphics[width=0.8\linewidth]{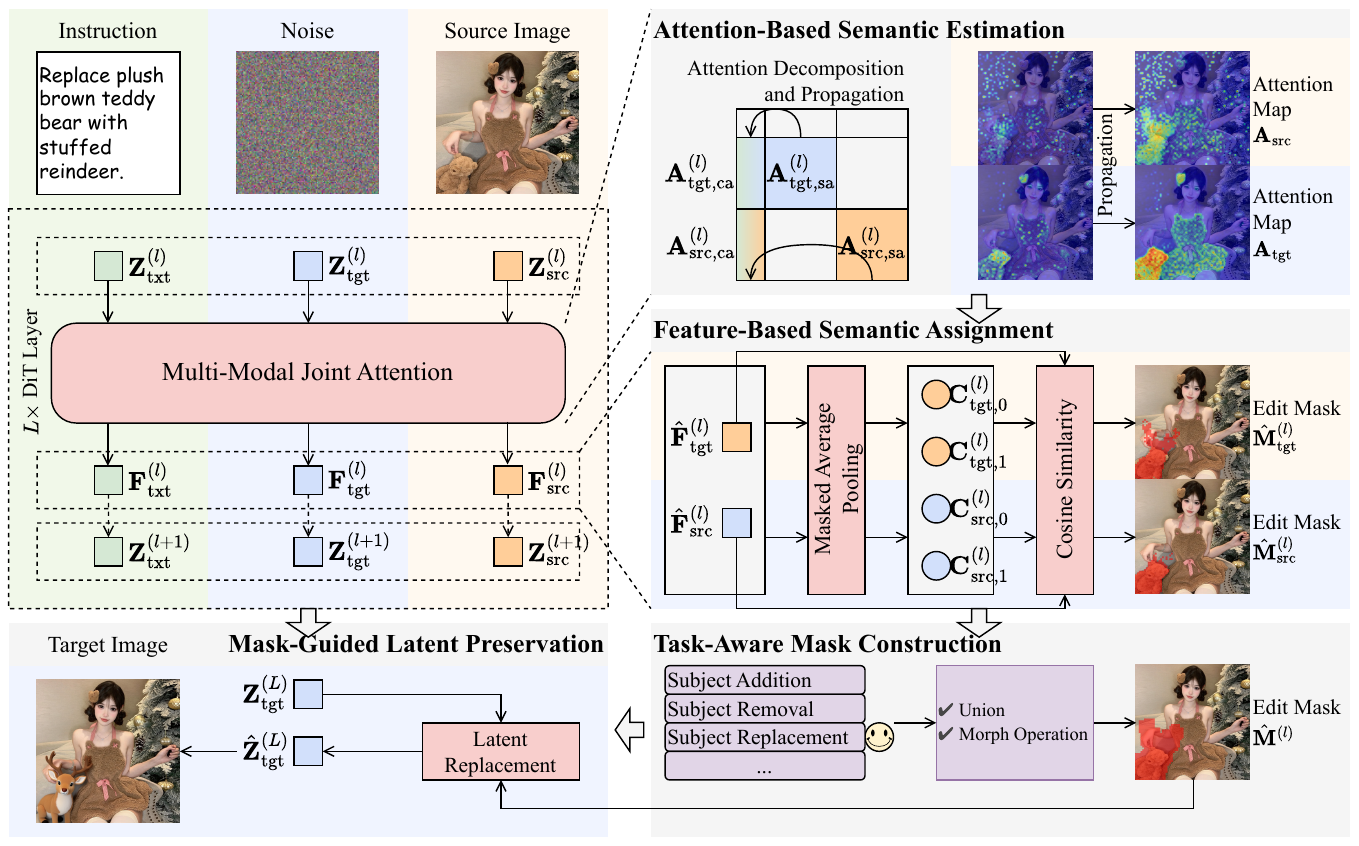}
    \caption{\small Overview of our framework. Given an editing instruction, an initial noise, and a source image, the model performs joint attention over text, target, and source tokens within each DiT layer. We first decompose and propagate attention to derive attention maps that provide coarse estimation of instruction-relevant regions (Attention-Based Semantic Estimation). The attention maps are leveraged to compute clustering centroids, and each token is assigned to the nearest centroid based on feature similarity (Feature-Based Semantic Assignment). We then construct the edit mask conditioned on the task type (Task-Aware Mask Construction). Finally, mask-guided latent preservation enforces localized edits while maintaining consistency in non-target regions (Mask-Guided Latent Preservation).}
    \label{fig:overview}
\end{figure*}

Recent instruction-based image editing models~\cite{liu2025step1x,wu2025qwen} are typically built upon diffusion transformers (DiTs)~\cite{peebles2023scalable} that operate over multi-modal latent representations, enabling joint modeling of the textual instruction, the source reference image, and the target edited image.
Specifically, the inputs to the DiT model consist of three types of tokens: text tokens $\mathbf{Z}_{\text{txt}} \in \mathbb{R}^{N_{\text{txt}} \times D}$, source image tokens $\textbf{Z}_{\text{src}} \in \mathbb{R}^{N_{\text{img}} \times D}$, and target image tokens $\mathbf{Z}_{\text{tgt}} \in \mathbb{R}^{N_{\text{img}} \times D}$, where $N_{\text{txt}}$ and $N_{\text{img}}$ denote the number of text and image tokens, respectively, and $D$ is the latent feature dimension.
The text and source tokens serve as static conditioning signals throughout the diffusion process, whereas the target tokens are iteratively updated at each denoising timestep under the formulation of rectified flow~\cite{liu2022flow}.
Concretely, the DiT model learns a continuous velocity field $v_{\theta}$ that transports the target tokens from an initial Gaussian distribution toward the data distribution.
The evolution of the target tokens can be described by the ordinary differential equation (ODE)~\cite{lipman2022flow}:
\begin{equation}
\label{eq:ode}
    \frac{d \mathbf{Z}_{\text{tgt}}(t)}{d t}=v_{\theta}\left(\mathbf{Z}_{\text{txt}}, \mathbf{Z}_{\text{tgt}}(t), \mathbf{Z}_{\text{src}}, t\right).
\end{equation}
To facilitate interaction across modalities, the DiT model employs multi-modal joint attention within each DiT layer, where tokens are concatenated along the token dimension to form a unified token sequence~\cite{tan2025ominicontrol}.
At denoising timestep $t$, the input token sequence to the attention module within the $l$-th DiT layer is denoted as $\mathbf{Z}^{(l)}(t) = \left [ \mathbf{Z}_{\text{txt}}^{(l)}; \mathbf{Z}_{\text{tgt}}^{(l)}(t); \mathbf{Z}_{\text{src}}^{(l)} \right ]$.
This unified token sequence is then projected into queries $\mathbf{Q}^{(l)}(t)$, keys $\mathbf{K}^{(l)}(t)$, and values $\mathbf{V}^{(l)}(t)$.
The multi-modal joint attention module is formulated as:
\begin{align}
    \mathbf{A}^{(l)}(t) &= \operatorname{Softmax} \left( \frac{\mathbf{Q}^{(l)}(t){\mathbf{K}^{(l)}(t)}^{\top}}{\sqrt{D}} \right), \label{eq:attn} \\
    \mathbf{F}^{(l)}(t) &= \mathbf{A}^{(l)}(t) \cdot \mathbf{V}^{(l)}(t), \label{eq:feat}
\end{align}
where $\mathbf{A}^{(l)}(t) \in \mathbb{R}^{(N_{\text{txt}}+2N_{\text{img}}) \times (N_{\text{txt}}+2N_{\text{img}})}$ is the joint attention matrix, and $\mathbf{F}^{(l)}(t) \in \mathbb{R}^{(N_{\text{txt}} + 2N_{\text{img}}) \times D}$ is the output latent feature of the joint attention module within the $l$-th DiT layer at timestep $t$.

\section{Method}
\label{sec:method}

Existing IIE models lack an explicit mechanism to spatially constrain latent updates.
While the learned velocity field $v_{\theta}$ captures global generative dynamics, the all-to-all coupling in the joint attention module, defined by the interaction among $\mathbf{Z}_{\text{txt}}$, $\mathbf{Z}_{\text{src}}$, and $\mathbf{Z}_{\text{tgt}}(t)$, lacks inherent spatial regularization.
Without an explicit inductive bias to partition the latent manifold, the denoising process treats all spatial tokens as equally mutable, which may lead to over-editing and degradation of instruction-irrelevant regions.

To enforce content consistency, we first construct task-aware edit masks to identify the subset of target tokens that requires updates in Sec.~\ref{sec:edit_localization}, and then introduce a preservation scheme that anchors the complementary subset to the source representation in Sec.~\ref{sec:latent_preservation}.
The overview of our framework is shown in Fig.~\ref{fig:overview}.

\subsection{Task-Aware Edit Localization}
\label{sec:edit_localization}

\subsubsection{Attention-Based Semantic Estimation.}
We first derive coarse edit localization by exploiting the joint attention matrix $\mathbf{A}^{(l)}(t)$ in Eq.~\eqref{eq:attn}.
Under the dual-stream image formulation, $\mathbf{A}^{(l)}(t)$ admits a block-wise structure, where submatrices encode interactions among text tokens, target tokens, and source tokens. 
As illustrated in the upper right panel of Fig.~\ref{fig:overview}, we selectively extract submatrices to capture (i) cross-modal interactions between text tokens and image tokens, and (ii) intra-modal interactions within each image stream.
Let $\mathcal{I}_{\text{txt}} = \{0, \dots, N_{\text{txt}}-1\}$, $\mathcal{I}_{\text{tgt}} = \{N_{\text{txt}}, \dots, N_{\text{txt}}+N_{\text{img}}-1\}$, and $\mathcal{I}_{\text{src}} = \{N_{\text{txt}}+N_{\text{img}}, \dots, N_{\text{txt}}+2N_{\text{img}}-1\}$ represent the discrete index sets for text tokens, target image tokens, and source image tokens, respectively.
For notational brevity, we use ``img'' $\in \{\text{``tgt''}, \text{``src''}\}$ to denote the image stream of interest.
The cross-attention submatrix $\mathbf{A}_{\text{img,ca}}^{(l)}(t) \in \mathbb{R}^{N_{\text{img}} \times N_{\text{txt}}}$ and the self-attention submatrix $\mathbf{A}_{\text{img,sa}}^{(l)}(t) \in \mathbb{R}^{N_{\text{img}} \times N_{\text{img}}}$ are obtained by slicing the joint attention matrix as follows:
\begin{equation}
\begin{cases}
    \mathbf{A}_{\text{img, ca}}^{(l)}(t) = \mathbf{A}^{(l)}(t)[\mathcal{I}_{\text{img}}, \mathcal{I}_{\text{txt}}], \\
    \mathbf{A}_{\text{img, sa}}^{(l)}(t) = \mathbf{A}^{(l)}(t)[\mathcal{I}_{\text{img}}, \mathcal{I}_{\text{img}}].
\end{cases}
\end{equation}
Drawing inspiration from the random-walk interpretation of attention on graphs~\cite{abnar2020quantifying,roffo2026infinite}, we treat the self-attention submatrix $\mathbf{A}_{\text{img,sa}}^{(l)}(t)$ as a transition matrix that captures token-wise affinities within the latent space.
Based on this interpretation, we perform a single-step diffusion of the linguistic signal to promote spatial coherence:
\begin{equation}
\label{eq:attn_propagation}
    \mathbf{A}_{\text{img}}^{(l)}(t) = \mathbf{A}_{\text{img,sa}}^{(l)}(t) \cdot \mathbf{A}_{\text{img,ca}}^{(l)}(t),
\end{equation}
where $\mathbf{A}_{\text{img}}^{(l)}(t) \in \mathbb{R}^{N_{\text{img}} \times N_{\text{txt}}}$ denotes the diffused attention.

Next, we aggregate the diffused attention $\mathbf{A}_{\text{img}}^{(l)}(t)$ across $L$ transformer layers and a subset of text tokens, followed by thresholding to obtain the attention-derived mask.
Let $\hat{\mathcal{I}}_{\text{txt}} \subseteq \mathcal{I}_{\text{txt}}$ denote the subset of text token indices that correspond to the semantically relevant part of the editing instruction.
In practice, this subset is automatically derived following the standard parsing procedures adopted in EdiVal-Bench~\cite{chen2025edival}.
The attention map $\mathbf{A}_{\text{img}}(t)$ and attention-derived mask $\mathbf{M}_{\text{img}}(t)$ are computed as:
\begin{align}
    \mathbf{A}_{\text{img}}(t) &= \operatorname{Norm}_{\text{min-max}} \left(
    % \frac{1}{L |\hat{\mathcal{I}}_{\text{txt}}|}
    \sum_{l}
    \sum_{i \in \hat{\mathcal{I}}_{\text{txt}}}
    \mathbf{A}_{\text{img}}^{(l)}(t)[:, i] \right), \\
    \mathbf{M}_{\text{img}}(t) &= \mathbf{A}_{\text{img}}(t) > \tau, \label{eq:attn_mask}
\end{align}
where $\tau \in (0.0, 1.0)$ is the attention threshold, and $\operatorname{Norm}_{\text{min-max}}(\cdot)$ denotes min-max normalization.
The two attention-derived masks, $\mathbf{M}_{\text{tgt}}(t)$ and $\mathbf{M}_{\text{src}}(t)$, provide distinct spatial priors that support edit localization of either synthesized or reference content.

\subsubsection{Feature-Based Semantic Assignment.}
Despite that the attention propagation defined in Eq.~\eqref{eq:attn_propagation} improves spatial coherence by considering token-wise affinities, attention-derived masks remain fundamentally limited in capturing semantic cues.
Specifically, attention primarily reflects instruction-conditioned relevance rather than structural semantics, and therefore may highlight tokens that are contextually related but not part of the target object.
Moreover, the absence of an explicit grouping mechanism in attention may lead to under- or over-inclusive regions, particularly near object boundaries~\cite{he2026early}.

To bridge this gap, as shown in the middle right panel of Fig.~\ref{fig:overview}, we treat the attention-derived masks as initial semantic indicators and perform a subsequent refinement in the latent feature space, where contextualized token representations produced by the joint attention module enable more precise and semantically coherent partitioning.
Recall that we define the attention-derived masks $\mathbf{M}_{\text{img}}(t)=\{\mathbf{M}_{\text{tgt}}(t), \mathbf{M}_{\text{src}}(t)\} \in \mathbb{R}^{N_{\text{img}}}$ in Eq.~\eqref{eq:attn_mask}, and the latent features $\mathbf{F}^{(l)}(t) \in \mathbb{R}^{(N_{\text{txt}}+2N_{\text{img}}) \times D}$ in Eq.~\eqref{eq:feat}.
For each image stream, we first extract and normalize the stream-specific latent features:
\begin{equation}
    \hat{\mathbf{F}}_{\text{img}}^{(l)}(t) = \operatorname{Norm}_{l_2} \left( \mathbf{F}^{(l)}(t)[\mathcal{I}_{\text{img}}, :] \right),
\end{equation}
where $\hat{\mathbf{F}}_{\text{img}}^{(l)}(t) \in \mathbb{R}^{N_{\text{img}} \times D}$ denotes $\ell_2$-normalized latent features for the image stream of interest.
The attention-derived mask $\mathbf{M}_{\text{img}}(t)$ is considered as a coarse partition that helps identify edited and preserved regions, from which we construct two feature centroids via masked average pooling.
Specifically, for each class $c \in \{0,1\}$, we compute:
\begin{equation}
    \mathbf{C}_{\text{img},c}^{(l)}(t) =
    \frac{\sum_{i} \hat{\mathbf{F}}_{\text{img}}^{(l)}(t)[i, :] \cdot \mathbb{I} \left( \mathbf{M}_{\text{img}}(t)=c \right)[i]}
    {\sum_{i} \mathbb{I} \left( \mathbf{M}_{\text{img}}(t)=c \right)[i] + \epsilon},
\end{equation}
where $\epsilon$ is a small constant for numerical stability.
The centroids $\mathbf{C}_{\text{img},1}^{(l)}(t) \in \mathbb{R}^{D}$ and $\mathbf{C}_{\text{img},0}^{(l)}(t) \in \mathbb{R}^{D}$ serve as representative features for the edited and preserved regions, respectively.
We then derive the edit mask by assigning each token to the closest centroid.
For each spatial token $i$, we compute its cosine similarity to both centroids and assign it according to:
\begin{equation}
\label{eq:feat_mask}
    \hat{\mathbf{M}}_{\text{img}}^{(l)}(t)[i] =
    \begin{cases}
        1 & \text{if } \hat{\mathbf{F}}_{\text{img}}^{(l)}(t)[i,:] \mathbf{C}_{\text{img},1}^{(l)}(t) > \hat{\mathbf{F}}_{\text{img}}^{(l)}(t)[i,:] \mathbf{C}_{\text{img},0}^{(l)}(t), \\
        0 & \text{otherwise}.
    \end{cases}
\end{equation}

\subsubsection{Task-Aware Mask Construction.}
Given the edit masks from two image streams, $\hat{\mathbf{M}}_{\text{tgt}}^{(l)}(t)$ and $\hat{\mathbf{M}}_{\text{src}}^{(l)}(t)$, a key remaining question is how to select or combine these spatial priors under different image editing scenarios.
We argue that an effective mask construction strategy is inherently task-dependent, as different image editing primitives induce semantic changes in different image streams.
Here we consider three representative subject-centric editing primitives: \textbf{subject addition}, \textbf{subject removal}, and \textbf{subject replacement}, which are distinguished by how semantic changes manifest across the source and target streams.
The task-aware edit mask $\hat{\mathbf{M}}^{(l)}(t)$ is derived according to the following rules:

\noindent \textbf{(a) Subject addition}: New content is introduced using the source image as a contextual prior for placement. However, as the added object is not present in the original source image, the localization signal is primarily manifested within the \emph{target} stream, i.e., $\hat{\mathbf{M}}^{(l)}(t)=\hat{\mathbf{M}}_{\text{tgt}}^{(l)}(t)$.

\noindent \textbf{(b) Subject removal}: Existing content is eliminated from the source image. The regions to be modified are therefore localized within the \emph{source} stream, i.e., $\hat{\mathbf{M}}^{(l)}(t)=\hat{\mathbf{M}}_{\text{src}}^{(l)}(t)$.

\noindent \textbf{(c) Subject replacement}: This task involves the simultaneous erasure of a source instance and the generation of a target instance, inducing coupled semantic changes that span both \emph{source} and \emph{target} streams, i.e., $\hat{\mathbf{M}}^{(l)}(t)=\hat{\mathbf{M}}_{\text{tgt}}^{(l)}(t) \cup \hat{\mathbf{M}}_{\text{src}}^{(l)}(t)$.

These principles can be readily generalized to other localized image editing tasks.
After obtaining the task-aware edit mask, we further apply a lightweight post-processing step to preserve the fidelity of the edited region.
Specifically, we retain the largest connected component to suppress spurious region and then fill small holes, followed by a moderate spatial expansion to ensure sufficient coverage of the edited region.

\begin{figure*}[htbp]
    \centering
    \includegraphics[width=1.0\linewidth]{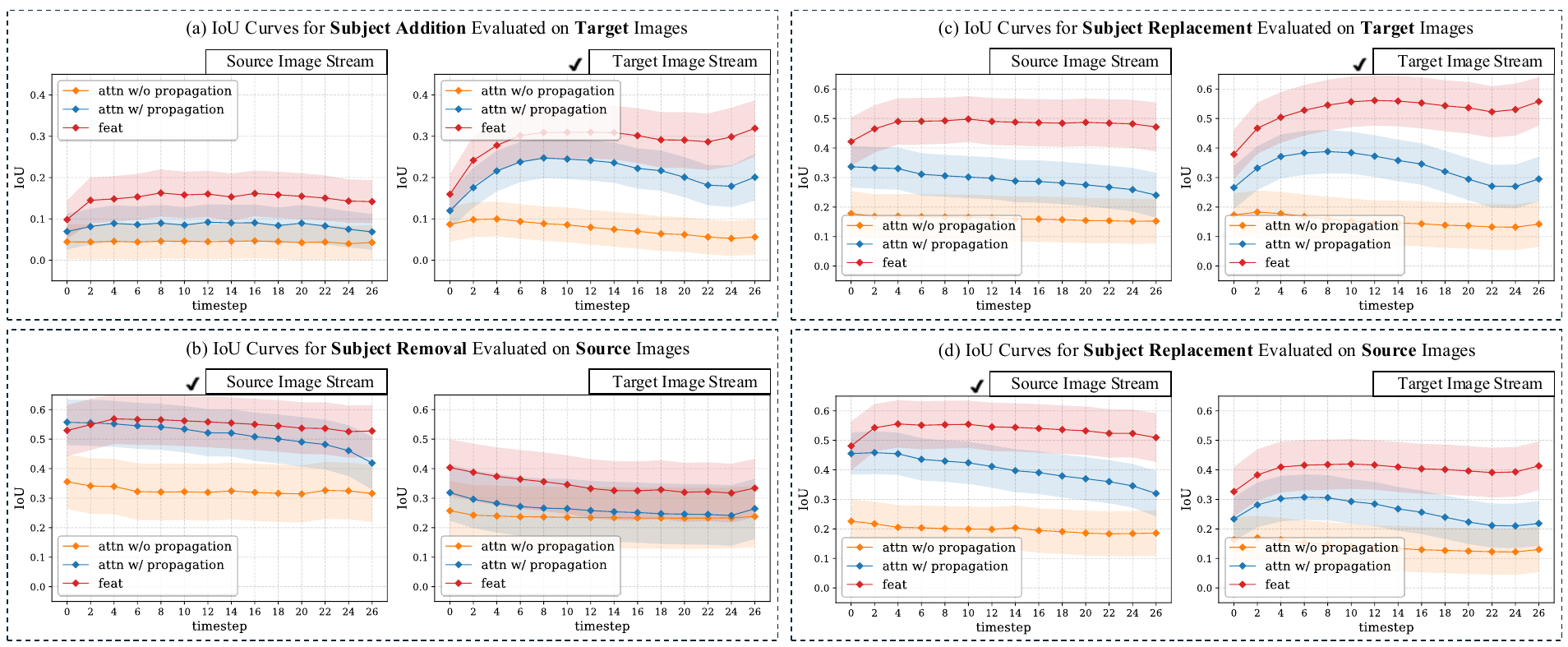}
    \caption{\small Quantitative results of edit localization analysis. There are two main observations: (i) The red curves generally achieve higher IoU than the orange and blue ones across denoising timesteps and task types, indicating that latent features provide stronger semantic cues. (ii) Semantic signals emerge from different image streams depending on the specific type of the editing task.}
    \label{fig:quantitative_analysis}
\end{figure*}

\begin{figure}[htbp]
    \centering
    \includegraphics[width=1.0\linewidth]{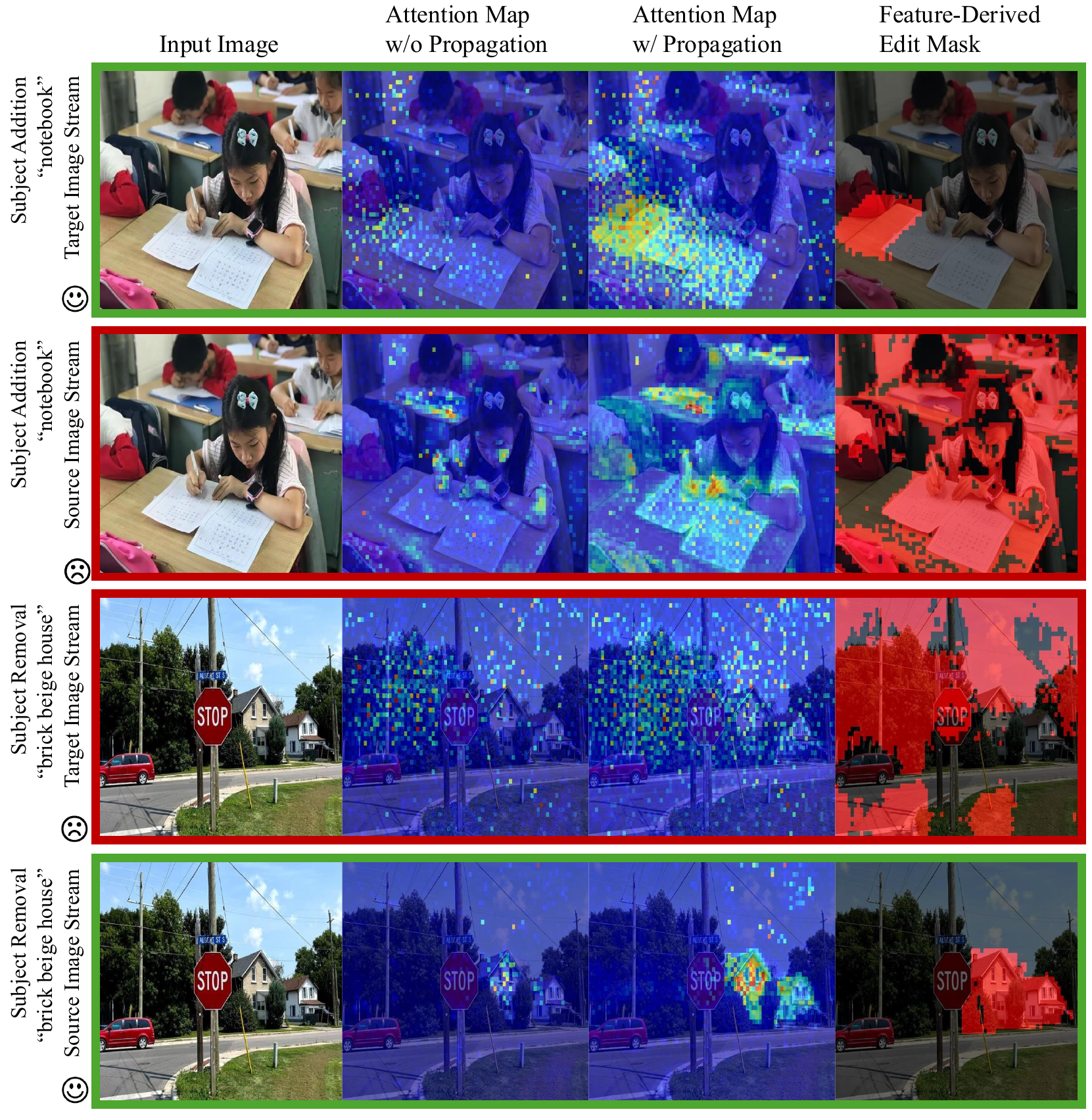}
    \caption{\small Qualitative results of edit localization analysis. (i) Feature-derived masks exhibit clearer boundaries and more complete spatial coverage than attention-derived ones. (ii) Edit semantics emerge from different image streams based on the task type.}
    \label{fig:qualitative_analysis}
\end{figure}

\subsection{Mask-Guided Latent Preservation}
\label{sec:latent_preservation}

Given the predicted edit mask $\hat{\mathbf{M}}^{(l)}(t)$, we now enforce spatially controlled image editing by constraining the evolution of target latents during the denoising process.
The key idea is to preserve the source-consistent region while allowing modifications only within the identified edit region.
Recall that the target image tokens evolve according to the learned dynamics defined in Eq.~\eqref{eq:ode}.
While the standard formulation updates all tokens uniformly, we introduce a mask-guided preservation mechanism that selectively constrains the update to within the identified edit region.
To achieve this, we construct an inverted latent that aligns with the current denoising timestep.
Specifically, let $\sigma_t \in [0,1]$ denote the noise schedule at timestep $t$, the inverted latent $\mathbf{Z}_{\text{inv}}(t) \in \mathbb{R}^{N_{\text{img}} \times D}$ is defined as:
\begin{equation}
\mathbf{Z}_{\text{inv}}(t) = \sigma_t \mathbf{Z}_{\text{tgt}}(0) + (1 - \sigma_t) \mathbf{Z}_{\text{src}}.
\end{equation}
Subsequently, we update the target image tokens by mask-guided latent blending:
\begin{equation}
\hat{\mathbf{Z}}_{\text{tgt}}(t) = \hat{\mathbf{M}}^{(l)}(t) \mathbf{Z}_{\text{tgt}}(t) + \left(1 - \hat{\mathbf{M}}^{(l)}(t)\right) \mathbf{Z}_{\text{inv}}(t).
\end{equation}
The DiT layer $l$ is set to a deep one to leverage high-level semantic signals for robust localization.
\section{Experiments}
\label{sec:experiments}

In this section, we first conduct a systematic analysis to empirically validate our edit localization framework using pseudo ground-truth annotations. We then compare our method with state-of-the-art image editing approaches, followed by ablation studies to assess the impact of key hyperparameters.

\subsection{Edit Localization Analysis}
\label{sec:analysis}

\subsubsection{Experimental Setup.}
We conduct the analysis across three representative editing primitives: subject addition, subject removal, and subject replacement.
We employ {Segment Anything Model (SAM) 3}~\cite{carion2025sam} to generate pseudo ground-truth segmentation masks, where relevant text phrases from the editing instructions are extracted as prompts to segment semantically matched instances.
For subject addition, we segment the object to be added on the generated target image.
For subject removal, we segment the object to be removed on the original source image.
For subject replacement, we perform both operations: segmenting the object to be added on the generated target image and the object to be removed on the source image.
We evaluate three types of localization masks: attention-based masks without propagation, attention-based masks with propagation ($\mathbf{M}_{\text{tgt}}(t)$ and $\mathbf{M}_{\text{src}}(t)$ in Eq.~\eqref{eq:attn_mask}), and feature-based masks ($\hat{\mathbf{M}}_{\text{tgt}}^{(l)}(t)$ and $\hat{\mathbf{M}}_{\text{src}}^{(l)}(t)$ in Eq.~\eqref{eq:feat_mask}).
For this analysis, we apply our framework to Qwen-Image-Edit~\cite{wu2025qwen}.
The attention threshold $\tau$ is set to $0.5$, and DiT layer $l$ is set to 50.
The number of denoising timesteps is fixed to $28$.
We employ average Intersection over Union (IoU) to evaluate the segmentation performance.

\subsubsection{Effectiveness of Feature-Based Semantic Assignment.}
The first observation is that latent features provide more reliable semantic cues for edit localization than attention-derived signals.
This is consistently reflected in Fig.~\ref{fig:quantitative_analysis}, where the red curves (feature-derived masks) generally achieve higher IoU scores than the orange and blue ones (attention-based masks) across denoising timesteps for three primitive editing tasks.
We note that the IoU scores for subject addition exhibit a noticeable drop compared to other tasks, which can be attributed to semantic ambiguity brought by subject addition instructions.
For instance, adding ``notebook'' to an image that already contains notebooks may cause the segmentation model to detect all instances, leading to less precise pseudo ground-truth masks.
Despite this, the actual instruction-following performance remains comparable to that of the original model, as shown in Tab.~\ref{tab:ablation_timestep}.

Quantitative examples in Fig.~\ref{fig:qualitative_analysis} further support these findings.
Attention-derived masks suffer from two main issues: noise and incomplete coverage.
For instance, in the subject addition case (1st row), attention maps are noisy and spill into irrelevant regions beyond the intended area of ``notebook''.
In the subject removal case (4th row), attention maps only capture the most salient part of ``brick beige house'' while missing its full extent.
In contrast, feature-derived masks exhibit cleaner boundaries and more complete spatial coverage, indicating stronger semantic consistency.

\subsubsection{Effectiveness of Task-Aware Mask Construction.}
The results in Fig.~\ref{fig:quantitative_analysis} also reveal that semantic signals for edit localization emerge from different image streams depending on the specific type of the editing task.
As illustrated, for subject addition, masks derived from the target image stream generally achieve higher IoU scores, indicating that addition-related semantics primarily manifest in the target stream.
Conversely, for subject removal, masks obtained from the source image stream exhibit superior performance, suggesting that removal-related semantics are strongly encoded within the source stream.
The same principle extends to subject replacement, where the source stream provides localization cues for the subject to be removed, while the target stream encodes the emerging semantics of the subject to be added.

This phenomenon is visually corroborated by the qualitative demonstrations in Fig.~\ref{fig:qualitative_analysis}.
For subject addition, a comparison of the first two rows shows that the ``notebook'' semantics clearly emerge in the target image stream.
Conversely, for subject removal, comparing the last two rows reveals that the ``brick beige house'' semantics are primarily captured in the source image stream.
These observations provide compelling empirical evidence for our proposed task-aware localization strategy.

\subsection{Comparison with State-of-the-arts}

\noindent \textbf{Baselines}.
We apply our edit localization framework to two state-of-the-art foundational image editing models: Step1X-Edit~\cite{liu2025step1x} and Qwen-Image-Edit~\cite{wu2025qwen}.
Both models employ a dual-stream architecture, where the target and source image tokens are concatenated along the token dimension and jointly processed within a diffusion transformer.
We compare our approach against a diverse set of instruction-based image editing methods, including InstructPix2Pix~\cite{brooks2023instructpix2pix}, MagicBrush~\cite{zhang2023magicbrush}, UltraEdit~\cite{zhao2024ultraedit}, ICEdit~\cite{zhang2025enabling}, and GRAG~\cite{zhang2025group}.
Among these, GRAG represents a strong training-free baseline that mitigates attention bias to achieve more precise and localized edits.

\noindent \textbf{Benchmark and Metrics.}
We evaluate all the methods on the recently introduced EdiVal-Bench~\cite{chen2025edival}, an \emph{object-centric} benchmark that aligns well with the goal of localized image editing.
% It yields strong agreement with human judgments in instruction following by using tools such as {object detector}~\cite{liu2024grounding} and {vision-language models}~\cite{qwen2}, which is more suitable for the image editing task.
The benchmark is built upon GEdit-Bench~\cite{liu2025step1x}, retaining 572 real-world images after filtering sensitive content.
The editing instructions are categorized into nine types: subject addition, subject removal, subject replacement, color alteration, material alteration, text change, position change, count change, and background change.
For the first three editing tasks, we directly apply the mask construction principles introduced in Sec.~\ref{sec:edit_localization}.
For other tasks, we derive edit regions from the source image stream for color alteration and material alteration, and from both streams for the remaining image editing tasks.

We adopt EdiVal-IF, EdiVal-CC, and EdiVal-O to evaluate instruction following, content consistency, and overall performance, respectively.
Particularly, content consistency is measured by averaging $l_1$ distance and DINO~\cite{caron2021emerging} feature similarity over non-target object and non-target background regions.
As a complementary metric, we employ Qwen-VL~\cite{qwen2.5} to rate the naturalness of edited images on a scale from 0 to 10 as perceptual quality, following~\cite{ku2024viescore}.

\noindent \textbf{Implementation Details.}
Input images are resized to approximately $1024 \times 1024$ while preserving the aspect ratio.
For all methods built on Step1X-Edit and Qwen-Image-Edit, the number of denoising timesteps is fixed to 28.
The attention threshold is set to $\tau=0.5$.
Due to differences in model depth, we set the feature extraction layer to $l=45$ for Step1X-Edit and $l=50$ for Qwen-Image-Edit.
The mask-guided latent preservation is applied progressively at timesteps 5, 10, and 15 to balance generation ability and content preservation.
We will discuss this choice in the following ablation study.
Our experiments are conducted on NVIDIA RTX A6000.
As our framework only relies on intermediate variables within the DiT model, the inference time and memory consumption are comparable to the original base models.

\subsubsection{Quantitative Results.}

\begin{table*}[tb]
    \centering
    \small
    \caption{\small Quantitative comparison results with state-of-the-art instruction-based image editing methods on the object-centric benchmark EdiVal~\cite{chen2025edival}. ``EdiVal-IF'', ``EdiVal-CC'', and ``EdiVal-O'' evaluate instruction following, content consistency, and the overall performance, respectively. We additionally employ a large vision-language model to rate the naturalness of edited images on a scale from 0 to 10 as perceptual quality.}
    \label{tab:quantitative_results}
    \begin{tabular}{llcccccc}
    \toprule
    \multirow{2.5}{*}{Method} & \multirow{2.5}{*}{Base Model} & \multirow{2.5}{*}{EdiVal-IF $\uparrow$} & \multicolumn{3}{c}{EdiVal-CC $\uparrow$} & \multirow{2.5}{*}{EdiVal-O $\uparrow$} & \multirow{2.5}{*}{\begin{tabular}[c]{@{}c@{}}Perceptual \\ Quality\end{tabular} $\uparrow$} \\
    \cmidrule(lr){4-6}
    & & & Object & Background & Overall & & \\
    \midrule
    InstructPix2Pix~\cite{brooks2023instructpix2pix} & SD 1.5      & 39.34 & 77.71 & 87.79 & 82.75 & 57.06 & 7.86 \\
    MagicBrush~\cite{zhang2023magicbrush}            & SD 1.5      & 42.66 & 82.90 & 93.62 & 88.26 & 61.36 & 7.89 \\
    UltraEdit~\cite{zhao2024ultraedit}               & SD 3        & 51.57 & 83.04 & 93.57 & 88.31 & 67.48 & 7.91 \\
    ICEdit~\cite{zhang2025enabling}                  & Flux.1 Fill & 53.50 & 88.21 & 93.92 & 91.07 & 69.80 & 7.91 \\
    \midrule
    Step1X-Edit~\cite{liu2025step1x} & Step1X-Edit & 59.09             & 90.73             & 97.32             & 94.03             & 74.54             & 7.86 \\
    +GRAG~\cite{zhang2025group}      & Step1X-Edit & 59.62~\dpos{0.53} & 91.63~\dpos{0.90} & 97.58~\dpos{0.26} & 94.60~\dpos{0.57} & 75.10~\dpos{0.56} & 7.90~\dpos{0.04} \\
    \rowcolor{rowgray}
    +Ours                            & Step1X-Edit & 60.84~\dpos{1.75} & 91.77~\dpos{1.04} & 97.80~\dpos{0.48} & 94.79~\dpos{0.76} & 75.94~\dpos{1.40} & 7.86~\dpos{0.00} \\
    \midrule
    Qwen-Image-Edit~\cite{wu2025qwen} & Qwen-Image-Edit & 70.80             & 86.51             & 93.78             & 90.14             & 79.89             & 7.96 \\
    +GRAG~\cite{zhang2025group}       & Qwen-Image-Edit & 67.31~\dneg{3.49} & 91.17~\dpos{4.66} & 95.49~\dpos{1.71} & 93.33~\dpos{3.19} & 79.26~\dneg{0.63} & 7.96~\dpos{0.00} \\
    \rowcolor{rowgray}
    +Ours                             & Qwen-Image-Edit & 71.15~\dpos{0.35} & 91.78~\dpos{5.27} & 96.96~\dpos{3.18} & 94.37~\dpos{4.23} & 81.94~\dpos{2.05} & 7.94~\dneg{0.02} \\
    \bottomrule
    \end{tabular}
\end{table*}

We present quantitative comparisons with state-of-the-art instruction-based image editing methods on EdiVal-Bench in Tab.~\ref{tab:quantitative_results}.
As can be seen, our framework consistently improves the performance of both foundational models across EdiVal-Bench metrics.
A key advantage lies in improving content consistency.
On Step1X-Edit, our method achieves gains in both object consistency (+1.04) and background consistency (+0.48), leading to a higher overall consistency score.
Notably, on Qwen-Image-Edit, our method significantly improves object consistency (+5.27) and background consistency (+3.18), substantially outperforming both the original base model~\cite{wu2025qwen} and the strong baseline GRAG~\cite{zhang2025group}.
These results validate that our task-aware localization and mask-guided preservation effectively reduce unintended modifications, leading to more faithful preservation of non-target regions.

Importantly, our improvements in content consistency do not come at the cost of instruction following.
The EdiVal-IF results on both models show that our method maintains, and in some cases improves, instruction-following capability when image editing is more focused.
In contrast, GRAG introduces a noticeable drop in instruction following on Qwen-Image-Edit (-3.49), suggesting a trade-off between content preservation and instruction adherence.

While our framework achieves comparable perceptual quality to the base model on Step1X-Edit, we observe a slight drop (0.02) on Qwen-Image-Edit.
This degradation primarily occurs in cases where the base model produces significant viewpoint changes.
Under such deviations, our mask-guided latent preservation method may enforce partial alignment with the original layout, leading to artifacts resembling image stitching.

\subsubsection{Qualitative Results.}

\begin{figure*}[tbp]
    \centering
    \includegraphics[width=0.9\linewidth]{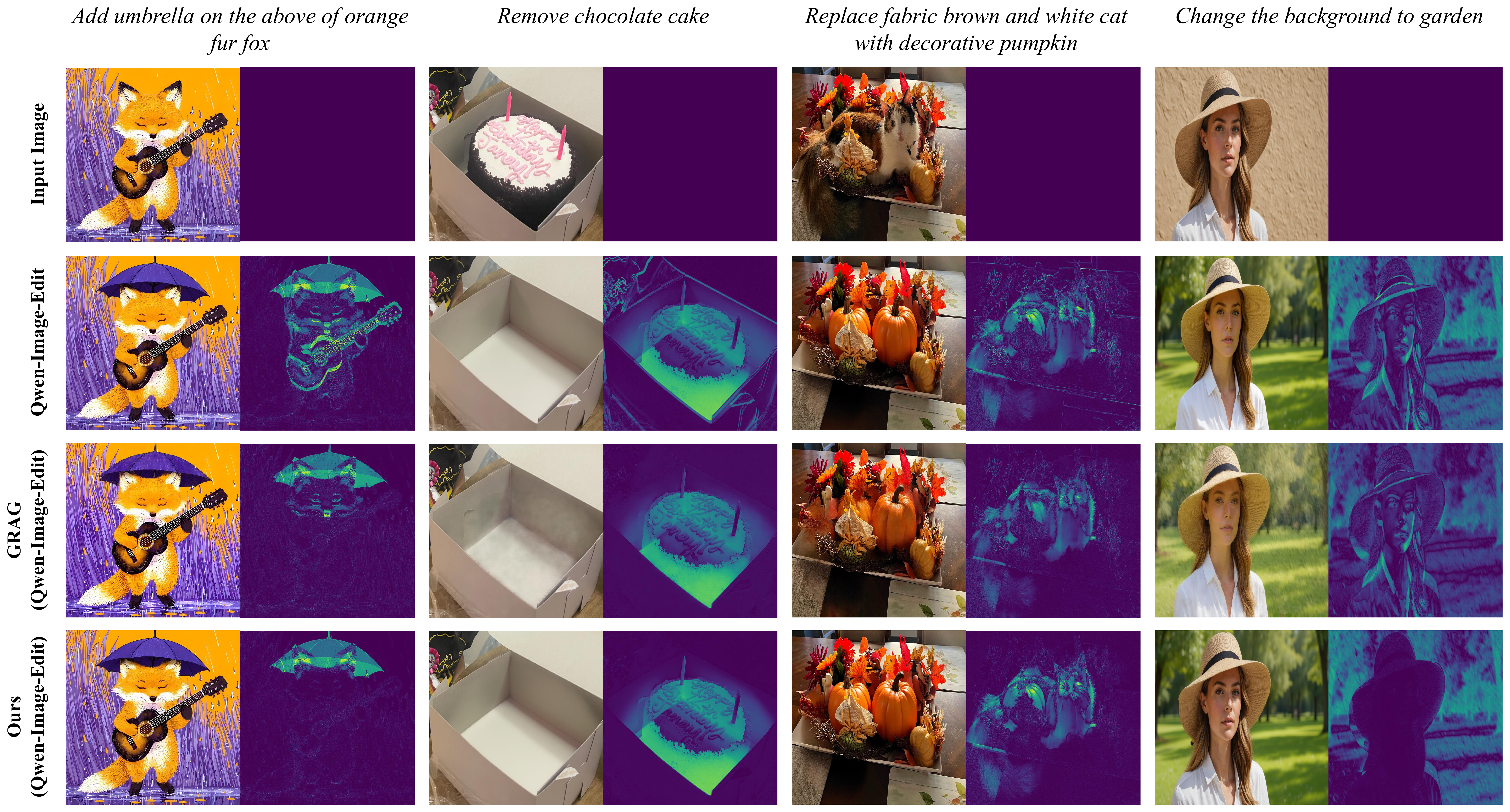}
    \caption{\small Qualitative comparisons with state-of-the-art instruction-based image editing methods. For each case, we display the edited image on the left and its corresponding pixel-wise difference map on the right. The results show that our method achieves superior content preservation while retaining the original editing capability of the base model.}
    \label{fig:qualitative_results}
\end{figure*}

We present qualitative comparisons on four editing instructions in Fig.~\ref{fig:qualitative_results}.
For each case, we display the edited image alongside its corresponding pixel-wise difference map, which highlights deviations from the input image.
% Overall, Step1X-Edit and Qwen-Image-Edit demonstrate robust instruction-following capabilities.
Here we present image editing results of methods built on Qwen-Image-Edit, as they exhibit the strongest instruction-following capability in Tab.~\ref{tab:quantitative_results}.
As we can see, the base model shows a tendency to modify unintended regions, as evidenced by the high-intensity areas in the difference maps.
For instance, it introduces noticeable changes to the original fox in the first example and alters the woman's appearance in the fourth example, despite these regions being unrelated to the editing instructions.
In contrast, our method preserves the generation quality of the base model while effectively maintaining the integrity of non-target regions, resulting in cleaner and more localized difference maps.
While GRAG also suppresses unintended edits through attention regulation, it occasionally compromises the visual fidelity of the synthesized content.
This is particularly evident in the second example, where the textural coherence of the cake box is visibly degraded.
These comparison results demonstrate that our framework achieves a superior balance between editing fidelity and content consistency.

\subsection{Ablation Study}

We conduct ablation studies on the attention threshold $\tau$, the DiT layer $l$, and the denoising timestep $t$ at which latent preservation is applied.
All ablations are performed on Qwen-Image-Edit.
% Unless ablated, the parameters are kept the same as in the comparison experiments.

\begin{figure}[tb]
    \centering
    \includegraphics[width=1.0\linewidth]{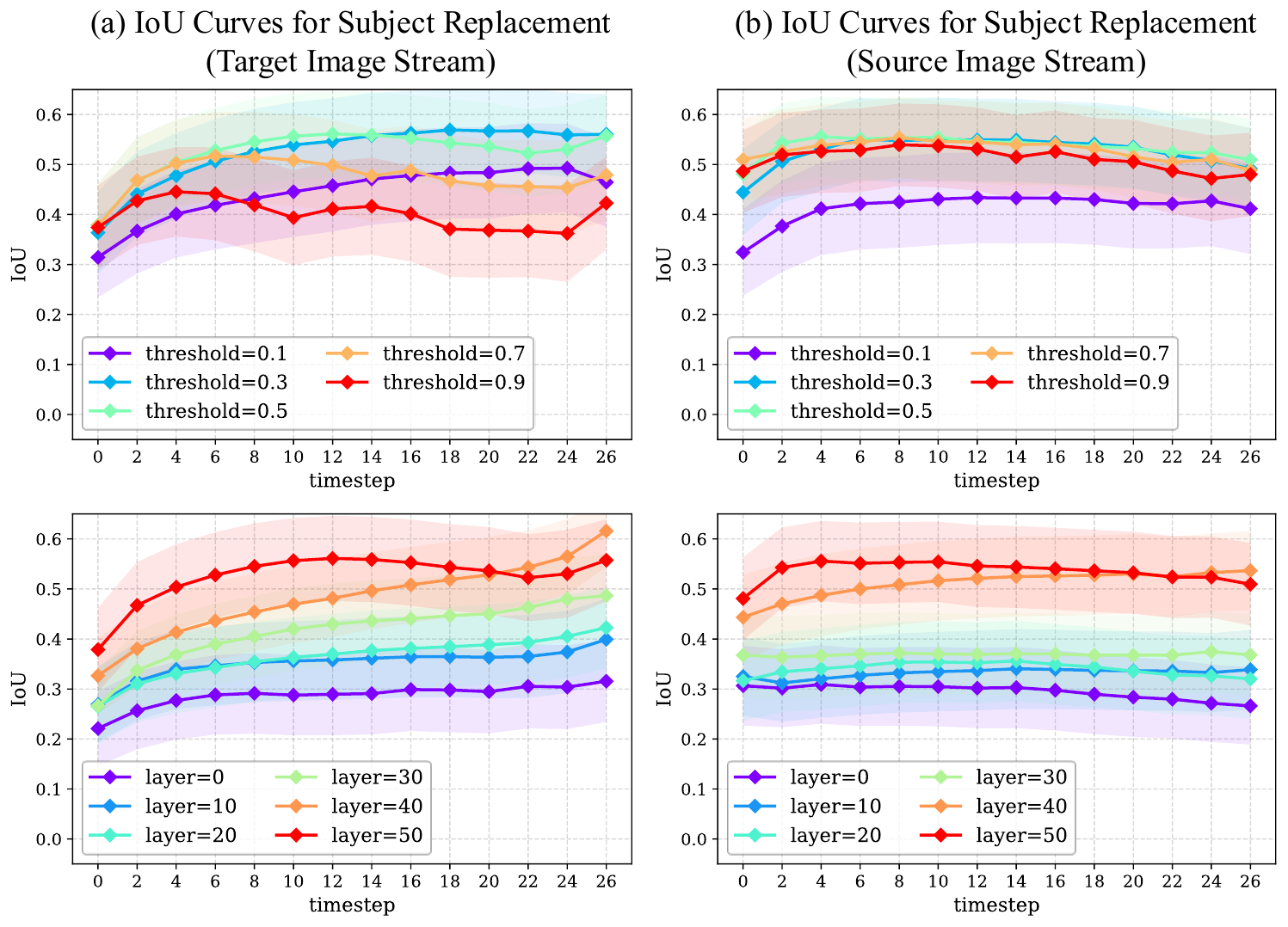}
    \caption{\small Ablation on attention threshold and DiT layer. (top) The thresholds ranging from 0.3 to 0.7 yield similar IoU scores, indicating that our method is robust to the choice of attention threshold. (bottom) The segmentation performance improves as the layer depth increases, with IoU scores peaking at layer 50 (red curves).}
    \label{fig:ablation}
\end{figure}

\subsubsection{Attention threshold.}

We examine the sensitivity of our framework to the attention threshold $\tau$ used for identifying clustering centroids.
As illustrated in Fig.~\ref{fig:ablation} (top), the thresholds ranging from $0.3$ to $0.7$ yield similar IoU scores for both source and target image streams, while extreme values (e.g., $0.1$ and $0.9$) lead to noticeable degradation.
This indicates that feature-based semantic assignment is robust to the choice of attention threshold.
We adopt a default attention threshold $\tau=0.5$ in our experiments.

\subsubsection{DiT layer.}

As illustrated in Fig.~\ref{fig:ablation} (bottom), we can observe a monotonic trend that segmentation performance improves as the layer depth increases, with IoU scores peaking at layer 50 (red curves).
This behavior suggests that edit semantics are progressively formed and become more discriminative in deeper layers of the diffusion transformer.
This finding contrasts with prior observation in semantic correspondence~\cite{gan2025unleashing} where intermediate layers are reported to be optimal.
We attribute this difference to the distinct requirements of the two tasks: semantic correspondence relies on high-level abstractions to match objects across images, whereas image editing requires precise instance-level grounding within a single image.
In deeper layers, features become highly specialized to local object boundaries and textures, making them particularly suited for semantic clustering.

\begin{table}[tb]
    \centering
    \small
    \caption{\small Ablation study on denoising timesteps for latent preservation. Each entry represents results for ``subject addition/subject removal/subject replacement''. ``PQ'' is short for perceptual quality.}
    \label{tab:ablation_timestep}
    \setlength{\tabcolsep}{2pt}
    \resizebox{\linewidth}{!}{\begin{tabular}{l cccc}
    \toprule
    Timesteps & EdiVal-IF $\uparrow$ & EdiVal-CC $\uparrow$ & EdiVal-O $\uparrow$ & PQ $\uparrow$ \\
    \midrule
    -               & 74.63/68.00/87.04 & 89.45/88.59/88.92 & 81.70/77.62/87.98 & 7.90/8.00/7.96 \\
    5               & 76.12/70.67/85.19 & 93.62/91.68/91.27 & 84.42/80.49/88.18 & 7.90/8.04/7.98 \\
    5, 10           & 76.12/70.67/85.19 & 94.28/92.44/91.86 & 84.71/80.82/88.46 & 7.88/7.99/8.00 \\
    \rowcolor{rowgray}
    5, 10, 15       & 77.61/70.67/85.19 & 94.53/92.94/92.02 & 85.65/81.04/88.54 & 7.88/7.97/7.98 \\
    5, 10, 15, 20   & 76.12/70.67/85.19 & 94.11/93.25/92.21 & 84.64/81.18/88.63 & 7.85/7.96/7.94 \\
    \bottomrule
    \end{tabular}}
\end{table}

\subsubsection{Denoising timestep.}
We ablate the denoising timestep at which mask-guided latent preservation is applied, with results summarized in Tab.~\ref{tab:ablation_timestep}.
Overall, EdiVal-IF remains largely insensitive to the choice of timesteps, indicating that constraining latent updates does not hinder the model's fundamental ability to follow editing instructions.
Notably, applying latent preservation as early as $t=5$ already yields a significant improvement in EdiVal-O.
As more timesteps are incorporated, EdiVal-CC continues to rise, reflecting increasingly strong preservation of non-edit regions.
However, as latent preservation is applied to more later denoising steps, we observe a gradual decline in naturalness scores.
This can be attributed to that excessively late-stage latent preservation may lead to a semantic mismatch, where the frozen source pixels fail to blend naturally with the newly synthesized content, resulting in visual artifacts near object boundaries.
Balancing these factors, we adopt $\{5, 10, 15\}$ as our default denoising timestep set.
\section{Conclusion}

In this work, we revisit instruction-based image editing through explicit edit localization and identify limited spatial control as a key contributor to over-editing in modern IIE models. Accordingly, we introduce a training-free, task-aware framework that explicitly identifies the regions to be edited. Our analysis shows that latent features provide stronger semantic grounding than attention maps and that effective localization is task-dependent. Experiments on EdiVal-Bench demonstrate that our method improves content consistency while maintaining instruction-following performance across diverse editing tasks. We hope this work encourages further exploration of internal representations in generative models for controlled visual synthesis.

\begin{acks}
This work was supported in part by the Australian Research Council under Projects DP240101848 and FT230100549.
\end{acks}

\bibliographystyle{ACM-Reference-Format}
\balance
\bibliography{main}

\end{document}